\documentclass[lettersize,journal]{IEEEtran}
\usepackage[implicit=false]{hyperref} 
\usepackage{amsmath,amsfonts}
\usepackage{algorithmic}
\usepackage{algorithm}
\usepackage{array}
\usepackage[caption=false,font=normalsize,labelfont=sf,textfont=sf]{subfig}
\usepackage{textcomp}
\usepackage{stfloats}
\usepackage{url}
\usepackage{verbatim}
\usepackage{graphicx}
\usepackage{cite}
\usepackage{multirow}
\usepackage{amsmath}
\usepackage{booktabs}
\usepackage{txfonts}
\usepackage{bbding}
\usepackage{pifont}
\usepackage{enumitem}
\begin{document}

\title{NavAgent: Multi-scale Urban Street View Fusion For UAV Embodied Vision-and-Language Navigation}

 \author{{Youzhi~Liu}, {Fanglong~Yao}\textsuperscript{*}\href{https://orcid.org/0000-0003-4187-9755}{\includegraphics[scale=0.07]{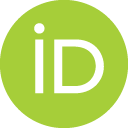}}, \IEEEmembership{Member,~IEEE,} {Yuanchang~Yue}, Guangluan Xu, \IEEEmembership{Member,~IEEE,} Xian Sun\href{https://orcid.org/0000-0002-0038-9816}{\includegraphics[scale=0.07]{figures/ORCIDiD_icon.png}}, \IEEEmembership{Senior~Member,~IEEE,} Kun~Fu, \IEEEmembership{Senior~Member,~IEEE}
   
\thanks{This work was supported by the National Natural Science Foundation of China under Grant 62306302. \textit{(Corresponding author: Fanglong~Yao)}} 
    \thanks{Youzhi Liu,~Yuanchang Yue,~Guangluan Xu,~Xian Sun,~Kun Fu are with the Aerospace Information Research Institute, Chinese Academy of Sciences, Beijing 100190, China, and also with the University of Chinese Academy of Sciences, Beijing 100190, China, and with the School of Electronic, Electrical and Communication Engineering, University of Chinese Academy of Sciences, Beijing 100190, China, and with the Key Laboratory of Network Information System Technology (NIST), Aerospace Information Research Institute, Chinese Academy of Sciences, Beijing 100190, China (e-mail:
   liuyouzhi22@mails.ucas.ac.cn;
   yueyuanchang22@mails.ucas.ac.cn;
   xugl@aircas.ac.cn;
    sunxian@aircas.ac.cn; kunfuiecas@gmail.com).}
    \thanks{Fanglong~Yao is with the Aerospace Information Research Institute, Chinese Academy of Sciences, Beijing 100190, China, and with the Key Laboratory of Network Information System Technology (NIST), Aerospace Information Research Institute, Chinese Academy of Sciences, Beijing 100190, China (e-mail: yaofanglong17@mails.ucas.ac.cn).}
	}

\markboth{Journal of \LaTeX\ Class Files,~Vol.~14, No.~8, August~2021}%
{Shell \MakeLowercase{\textit{et al.}}: A Sample Article Using IEEEtran.cls for IEEE Journals}


\maketitle

\begin{abstract}
Vision-and-Language Navigation (VLN), as a widely discussed research direction in embodied intelligence, aims to enable embodied agents to navigate in complicated visual environments through natural language commands. Most existing VLN methods focus on indoor ground robot scenarios. However, when applied to UAV VLN in outdoor urban scenes, it faces two significant challenges. First, urban scenes contain numerous objects, which makes it challenging to match fine-grained landmarks in images with complex textual descriptions of these landmarks. Second, overall environmental information encompasses multiple modal dimensions, and the diversity of representations significantly increases the complexity of the encoding process. To address these challenges, we propose NavAgent, the first urban UAV embodied navigation model driven by a large Vision-Language Model. NavAgent undertakes navigation tasks by synthesizing multi-scale environmental information, including topological maps (global), panoramas (medium), and fine-grained landmarks (local). Specifically, we utilize GLIP to build a visual recognizer for landmark capable of identifying and linguisticizing fine-grained landmarks. Subsequently, we develop dynamically growing scene topology map that integrate environmental information and employ Graph Convolutional Networks to encode global environmental data. In addition, to train the visual recognizer for landmark, we develop NavAgent-Landmark2K, the first fine-grained landmark dataset for real urban street scenes. In experiments conducted on the Touchdown and Map2seq datasets, NavAgent outperforms strong baseline models. The code and dataset will be released to the community to facilitate the exploration and development of outdoor VLN.
\end{abstract}

\begin{IEEEkeywords}
UAV Vision-and-Language Navigation, Large language models, Topology map.
\end{IEEEkeywords}

\section{Introduction}
\IEEEPARstart{U}{AV} Vision-and-Language Navigation (VLN) is a specialization of embodied intelligence for navigation applications in the aerial domains\cite{Liu2023AerialVLNVN,2017Vision,2020REVERIE,2019Stay,2020Room,2020Beyond,Chen2018TOUCHDOWNNL}. It aims to explore how to enable UAV embodied agents to navigate in unknown urban environments based on natural language commands and environmental observations. This approach has a wide range of applications across various fields, including inspection and monitoring, search and rescue, and low-altitude logistics\cite{Dji1,Dji2,Dji3}. However, existing research on VLN primarily focuses on indoor scenes\cite{10006384,Lin_2024,10359152,an2024etpnavevolvingtopologicalplanning,10120966}. In contrast, the outdoor urban environments targeted by UAV VLN tasks involve a much larger spatial scale, greater complexity, and sparser landmarks, making them more challenging\cite{0A,2010Vision,2018Mapping,2018Following,2018Map}.

Recent studies have demonstrated that large Vision-Language Models, which are capable of processing multimodal inputs, exhibit strong generalization abilities and outstanding performance in embodied tasks\cite{Li2023BLIP2BL,liu2023visualinstructiontuning,2021Grounded,2023InstructBLIP,peng2023kosmos2groundingmultimodallarge,zhang2024llamaadapterefficientfinetuninglanguage,Zhu2023MiniGPT4EV,Bai2023QwenTR}. Given this, our objective is to develop the first embodied large Vision-Language Model specifically designed for UAV VLN tasks, enabling UAV agents to navigate autonomously in urban street scenes. However, there are two primary challenges in this process.

(1) \textbf{Difficulty in Matching Fine-Grained Landmarks in Panoramic Observation Images.} When the agent is positioned at any observation point, it perceives the surrounding environment through a panoramic image captured at that location. The landmarks that need to be recognized are typically fine-grained targets located on both sides of the road, which comprise less than $5\%$ of the pixels in the panoramic image. Furthermore, the texts associated with these landmarks are often not simple nouns but rather complex phrases that include multiple modifiers, such as “a green mailbox” or “two red garbage cans”. As a result, ordinary image encoders struggle to accurately match these intricate details.

(2) \textbf{Difficulty in Encoding Overall Environmental Information in the Decision-Making Process.} The environments in which the agents operate are complex, requiring the integration of various dimensions of overall environmental information. This includes visual data (e.g., observation images), semantic information (e.g., landmark categories and locations), and geographic data (e.g., environmental map). Not only do these data types have different representations, but they also exhibit a high degree of heterogeneity in both space and time, which complicates the encoding process. Furthermore, the dynamic nature of the environmental information necessitates real-time updates as the agent moves, significantly increasing the challenges associated with coding.

\begin{figure*}[t]
	\centering
		\includegraphics[width=\textwidth]{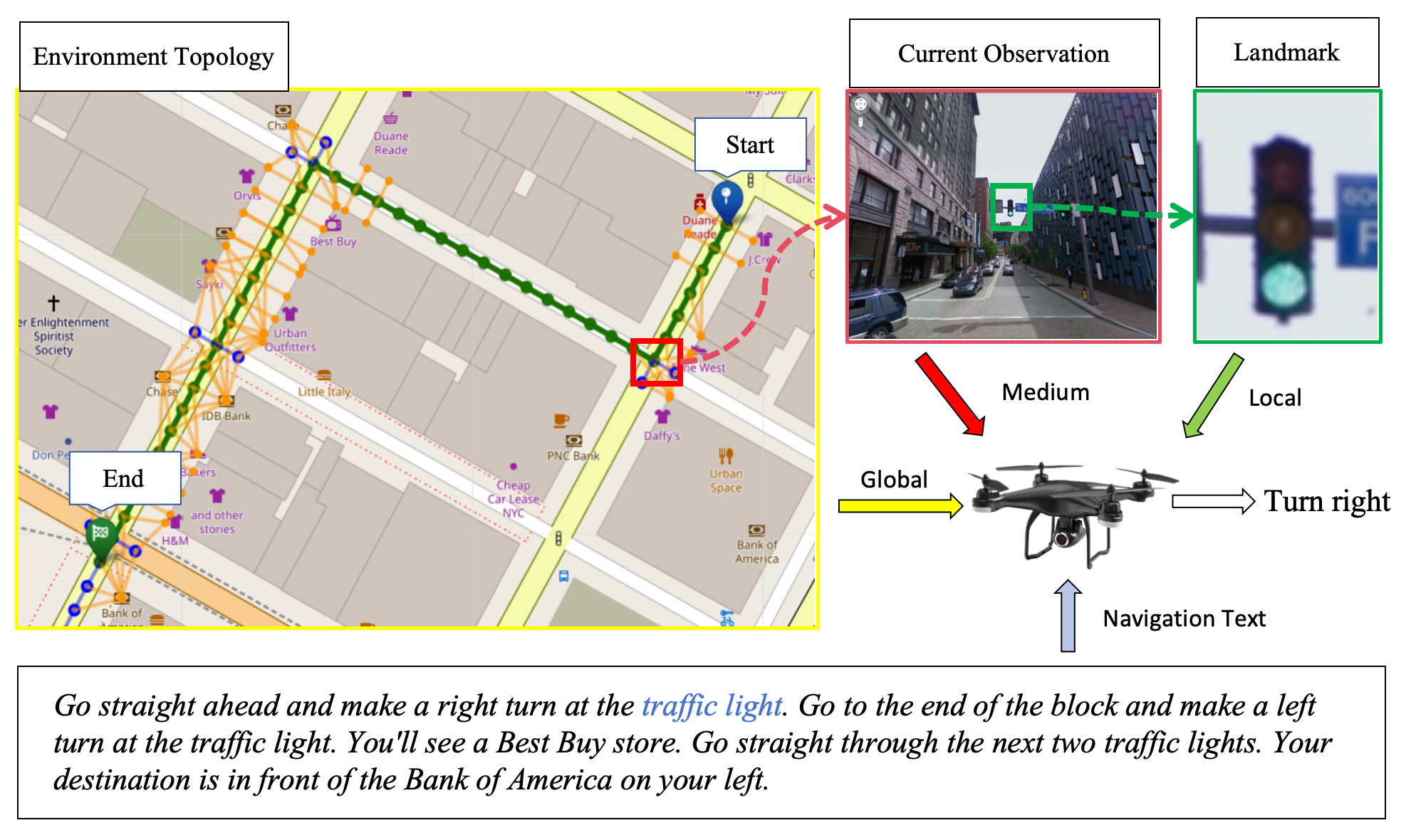}
	\caption{Schematic diagram of the VLN model augmented by multi-scale environment fusion, with the environment topology map containing the overall information of the environment in the yellow box, the observation image of the agent at this point in the red box, the fine-grained landmarks extracted from the observation image in the green box, and the navigation text in the black box .
 }
	\label{FIG:framework}
\end{figure*}

To address the aforementioned challenges, we propose a multi-scale environment fusion-enhanced VLN model called NavAgent. As illustrated in Figure \ref{FIG:framework}, this model integrates the global environmental topology map, the panoramic view of the current observation position, and local fine-grained landmark data. This integration facilitates accurate and stable VLN for UAV agents, specifically:

\renewcommand{\dblfloatpagefraction}{0.9}

Prior research has demonstrated the effectiveness of GLIP in fine-grained target recognition and matching tasks within a general-purpose domain\cite{2021Grounded}. 
We modify the structure of GLIP to develop the visual recognizer for landmark. The visual recognizer facilitates fine-grained matching between the images of the environment observed by the agent during navigation and the text of landmarks, enabling precise identification of landmarks present in the observation images.
To train the visual recognizer for landmark, we first frame the landmarks within the street images selected from Google Street View\cite{Ali_bey_2022}. We then use BLIP2\cite{Li2023BLIP2BL} to generate descriptions for the images of the landmarks, ultimately creating a dataset with 2,000 landmark annotations. This dataset, named NavAgent-Landmark2K, is the first landmark recognition dataset designed for outdoor VLN. Comparative experiments have shown that the visual recognizer for landmark fine-tuned with this dataset improves accuracy by $9.5\%$ compared to the GLIP in recognizing landmark images within the context of outdoor VLN.

Further, we develop a dynamically evolving scene topology map to integrate environmental information and design the topology map encoder to capture global environmental features. Specifically, we record navigable positions in the urban scene as nodes, initially capturing each node's position and the orientation relationships between nodes. We then explore the current node and its contiguous nodes, combining them into a cohesive scene topology map. To integrate environmental information, we employ an image encoder to extract visual features from the current observation image. We then utilize a cross-attention mechanism to incorporate these visual features into the scene topology map, facilitating the fusion of multimodal information. After the UAV agent moves, the scene topology map retains historical information and updates the nodes to integrate new environmental data, effectively encoding dynamic and complex environments.

In summary, our main contributions are:

(1) We propose NavAgent, the first urban UAV embodied navigation model driven by a large Vision-Language Model, enabling autonomous navigation of agent in urban environments through the fusion of multi-scale environmental information.

(2) We design and train the visual recognizer for landmark that recognizes fine-grained landmarks by calculating similarity scores by matching region features extracted from observed images with text features extracted from landmark descriptions. Experimental results indicate that the visual recognizer of landmark enhances accuracy by 9.5\% in the fine-grained landmark recognition task when compared to the GLIP.

(3) We construct a dynamically growing scene topology map and employ the topology map encoder to encode individual nodes and their spatial relationships, thereby enhancing the planning ability of agent for long-distance navigation.

(4) We develop the first fine-grained landmark dataset for real urban street scenes, named NavAgent-Landmark2K. This dataset comprises 2,000 image-text pairs, where the images represent fine-grained landmarks occupying approximately $5\%$ of the pixel area, and the accompanying text consists of landmark phrases that include multiple modifiers.

(5) In the experiments conducted on the Touchdown and Map2seq datasets, our proposed NavAgent outperforms the powerful baseline models, achieving improvements of $4.6\%$ and $2.2\%$ compared to VELMA on the development sets of two datasets.

\section{Related Work}
\subsection{Outdoor VLN}
In the outdoor VLN task\cite{Liu2023AerialVLNVN,Chen2018TOUCHDOWNNL,2019The,2020Learning,2020Retouchdown,2021SILG,Sun2023OutdoorVN,armitage2022prioritymapvisionandlanguagenavigation}, the agent must navigate in complex urban environments to reach target points based on human commands. Therefore, the ability of the agent to process and analyze the complex environments is crucial for accurate navigation. Hermann et al.\cite{2020Learning} extract features from observation images using networks such as CNNs and then fuse these features with textual data into a classifier, thereby generating navigation decisions. Zhu et al.\cite{2021Multimodal} utilize pre-trained vision and language transformers as a foundation, subsequently fine-tuning them for a task-specific dataset. Shah et al.\cite{shah2022lmnavroboticnavigationlarge} employ the CLIP model to assign scores for the presence of landmarks in images at each observation node, allowing for route planning based on this landmark labeling information. Schumann et al.\cite{2023VELMA} verbalize visual observations through a pipeline, enabling the agent to make decisions based solely on the current node environment. In contrast, our embodied agent utilizes environmental topology map to store historical data and integrate current information, enabling decision-making at multiple scales without relying on any prior knowledge of the environment.

\subsection{Application of LLM in embodied navigation}
Large language models (LLMs), such as OPT\cite{2022OPT}, PaLM\cite{Driess2023PaLMEAE}, ChatGPT\cite{2022Training}, and GPT-4\cite{openai2024gpt4technicalreport}, have demonstrated remarkable capabilities across various domains\cite{Touvron2023LLaMAOA}. Whereby training on extensive corpora, LLMs exhibit excellent planning and reasoning abilities. In the field of embodied navigation, Zhou et al.\cite{zhou2023escexplorationsoftcommonsense} utilize LLMs to extract common-sense knowledge about target-object relationships in observations for zero-shot target navigation. Zhou et al.\cite{zhou2023navgptexplicitreasoningvisionandlanguage} propose a command-tracking navigation agent, NavGPT, which is entirely based on the LLM (GPT-4) to validate zero-shot sequential action prediction for VLN, leveraging the reasoning abilities of the GPT model in complex embodied scenarios. Additionally, Shah et al.\cite{shah2022lmnavroboticnavigationlarge} introduce LM-Nav, a system for robot navigation that integrates an LLM (GPT-3), a large Vision-Language Model (CLIP), and a visual navigation model (ViNG) to facilitate long-distance navigation in unknown environments. Lin et al.\cite{Lin_2024} introduce a novel correctable landmark discovery agent that leverages two large-scale models, ChatGPT and CLIP, to implement a correctable landmark discovery scheme. This approach treats VLN as an open-world sequential landmark discovery problem. In contrast, our embodied agent, NavAgent, applies large Vision-Language Models for the first time to a UAV navigation task in an urban neighborhood.

\subsection{Application of Maps in embodied navigation}
Maps play an important role in the field of navigation as they provide an effective spatial representation\cite{2017Neural,2021EgoMap,2020Semantic,2024Audio,Joa2018MapNet,an2023bevbertmultimodalmappretraining}. Gupta et al.\cite{2017Cognitive} utilize a differentiable neural network planner to determine the next action at each time step. Meanwhile, Cartillier et al.\cite{2021Semantic} develop an egocentric semantic mapping network that leverages RGB-D observations and employ an encoder-decoder architecture to extract features. Georgakis et al.\cite{2022Cross} employ a cross-modal attention mechanism to learn map semantics and subsequently predict paths to a goal in the form of a set of waypoints. Chen et al.\cite{2021Topological} propose a modular VLN approach that utilizes topological map. Given natural language instructions and a topological map, an attentional mechanism predicts the navigation plan within the map. However, their proposed topological map is constructed in advance through environmental exploration, which means that the agent has access to global a priori topological information during navigation. This reliance on a pre-constructed map limits the approach's applicability in unfamiliar scenarios. In contrast, our embodied agent, NavAgent, enhances the topology map by incorporating updates from the visited environment, effectively capturing the layout of the environment. Additionally, we design a topology map encoder to extract environmental information, which addresses the issue of nodes being independent and lacking mutual attention. This is achieved by fusing the features of each node with those of other nodes.

\section{Datasets}
\subsection{Urban VLN Environment}
The environment used in this experiment is the Touchdown environment proposed by Chen et al\cite{Chen2018TOUCHDOWNNL}. It consists of a Google Street View representation of the Manhattan area in New York City, comprising 29,641 panoramic images connected by directed graphs \( G = \langle V, E \rangle \). Each directed graph contains multiple navigable points \( v \in V \) and edges \( \langle v, v' \rangle \in E \) that connect pairs of navigable points. The state of the agent at each navigable point can be represented as \( s = \langle v, \alpha \rangle \), where \( \alpha \) represents the heading from node \( v \) to node \( v' \). The action space available to the agent at each navigable point is \( \{ \text{FORWARD}, \text{LEFT}, \text{RIGHT}, \text{STOP} \} \).


\subsection{VLN datasets}
\begin{figure*}[t]
	\centering
		\includegraphics[width=\textwidth]{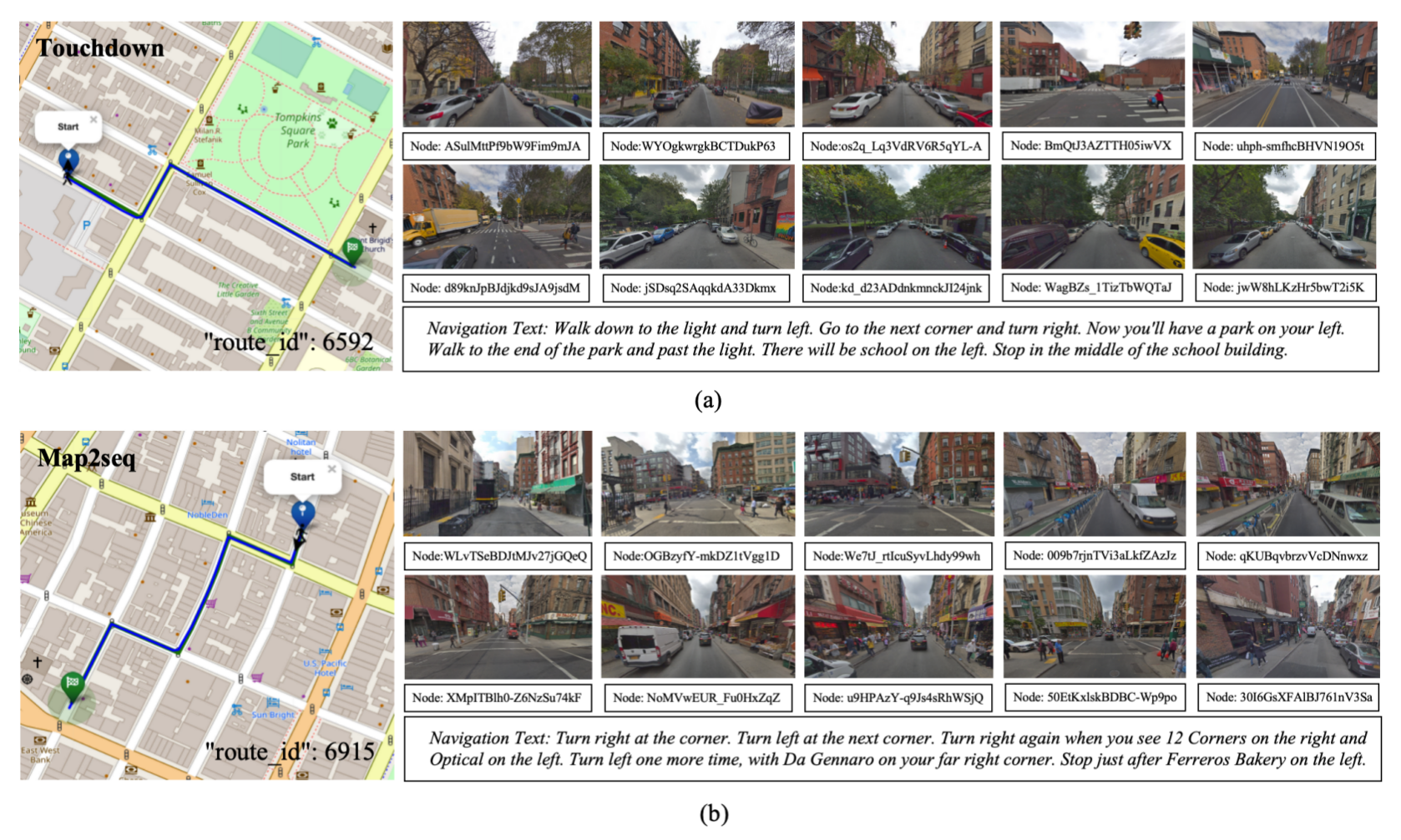}

	\caption{Examples of the Touchdown and Map2seq datasets. In Fig. (a), an example of the Touchdown dataset is presented, featuring the navigated gold route on the left, several nodes along the route with their corresponding observation images displayed above, and the navigation text at the bottom. An example of the Map2seq dataset is shown in Fig. (b), maintaining the same layout as in Fig. (a).
 }
	\label{FIG:dataset}
\end{figure*}

The datasets used in this experiment are the Touchdown\cite{Chen2018TOUCHDOWNNL} and Map2seq\cite{2020Generating} datasets, both of which are VLN instruction datasets based on navigation paths in the Touchdown environment. The Touchdown dataset comprises text descriptions of navigation instructions created by annotators based on predefined routes within the environment, along with panoramic images documenting the navigation process. It contains a total of 18,402 navigation instances. The Map2seq dataset, on the other hand, comprises navigation description texts created by annotators based solely on the navigation routes and landmark descriptions in the map, containing a total of 15,009 navigation instances. Both datasets include seen and unseen environments, which are categorized into training, development, and test sets, with the exact numbers provided in Table \ref{dataset}. In addition, in the Touchdown dataset, the initial orientation of the agent is random, while in the Map2seq dataset, it is aligned with the correct direction. An example of two datasets is presented in Figure \ref{FIG:dataset}.
\renewcommand{\dblfloatpagefraction}{0.9}

\begin{table}[hbt]
\centering
\caption{Data distribution in the Touchdown and Map2seq datasets.}
\label{dataset}
\begin{tabular}{c|ccc|ccc}
\toprule
\phantom{} & \multicolumn{3}{c|}{\centering Touchdown} & \multicolumn{3}{c}{\centering Map2seq} \\
\midrule
\phantom{} & train & dev & test & train & dev & test \\
\midrule
seen & 6,525 & 1,391 & 1,409 & 6,072 & 800 & 800\\
unseen & 6,770 & 800 & 1,507 & 5,737 & 800 & 800\\
\midrule
merged & 13,295 & 2,191 & 2,916 & 11,809 & 1,600 & 1,600\\
\bottomrule
\end{tabular}
\end{table}
\subsection{NavAgent-Landmark2K dataset}
\renewcommand{\dblfloatpagefraction}{0.9}
\begin{figure}[h]
	\centering
		\includegraphics[scale=.4]{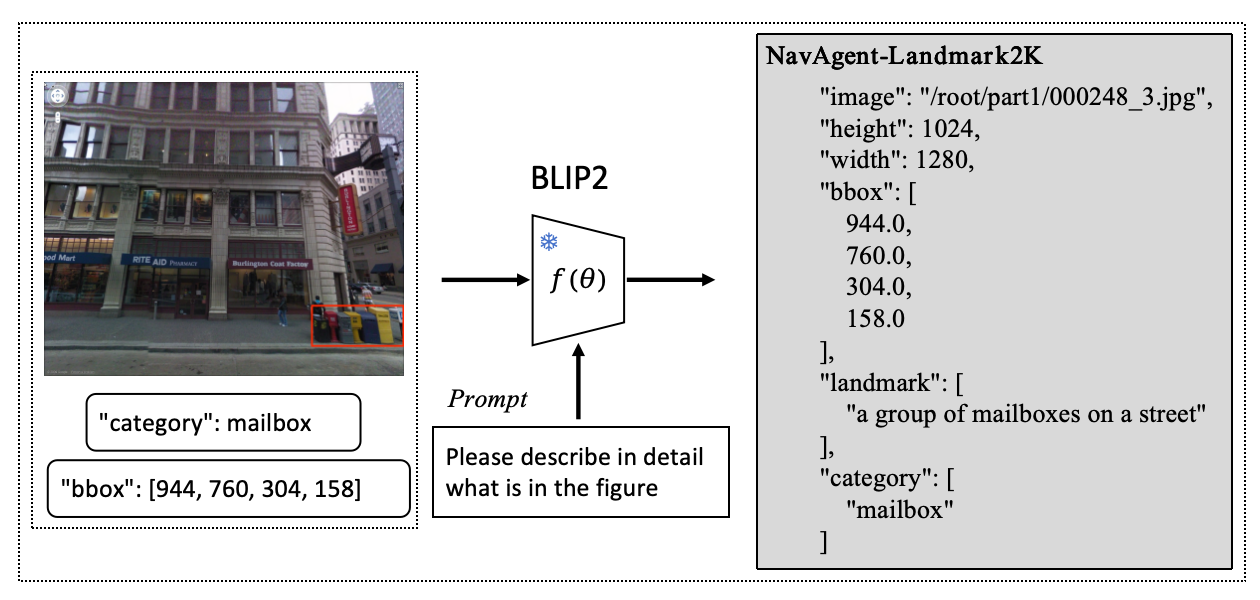}
	\caption{Figure shows the construction process and a specific example of the NavAgent-Landmark2K dataset. 
 }
	\label{FIG:gen1}
\end{figure}
In the process of VLN, enhancing the model's ability to recognize landmarks at the phrase level requires specific training for fine-grained target detection in the observed images. The GLIP model leverages large-scale data to learn language-aware and semantically enriched visual representations at the object level, demonstrating strong performance in the general domain, so we choose GLIP as the base model for detecting fine-grained landmarks. To train the detection model for the specialized domain, we construct the first fine-grained landmark dataset for real urban street scenes. and utilize this dataset for fine-tuning. Specifically, based on the GSV dataset\cite{Ali_bey_2022} from Google Street View, we obtain image data with landmark bounding boxes by having annotators outline common landmarks in the urban street view images, while also recording the types of landmarks. These images are subsequently processed using the BLIP2 model\cite{Li2023BLIP2BL}, which demonstrates strong image comprehension abilities in zero-shot scenarios. Before captioning, the images are cropped according to the bounding boxes to ensure that the generated captions accurately reflect the locations of the landmarks. The dataset is created by establishing a one-to-one correspondence among the images, bounding boxes (bbox), and captions\cite{lin2015microsoftcococommonobjects}. The construction process is illustrated in Figure \ref{FIG:gen1}, while an example of the dataset is shown on the right of Figure \ref{FIG:gen1}. 
\renewcommand{\dblfloatpagefraction}{0.9}
\begin{figure*}[t]
	\centering
		\includegraphics[width=\textwidth]{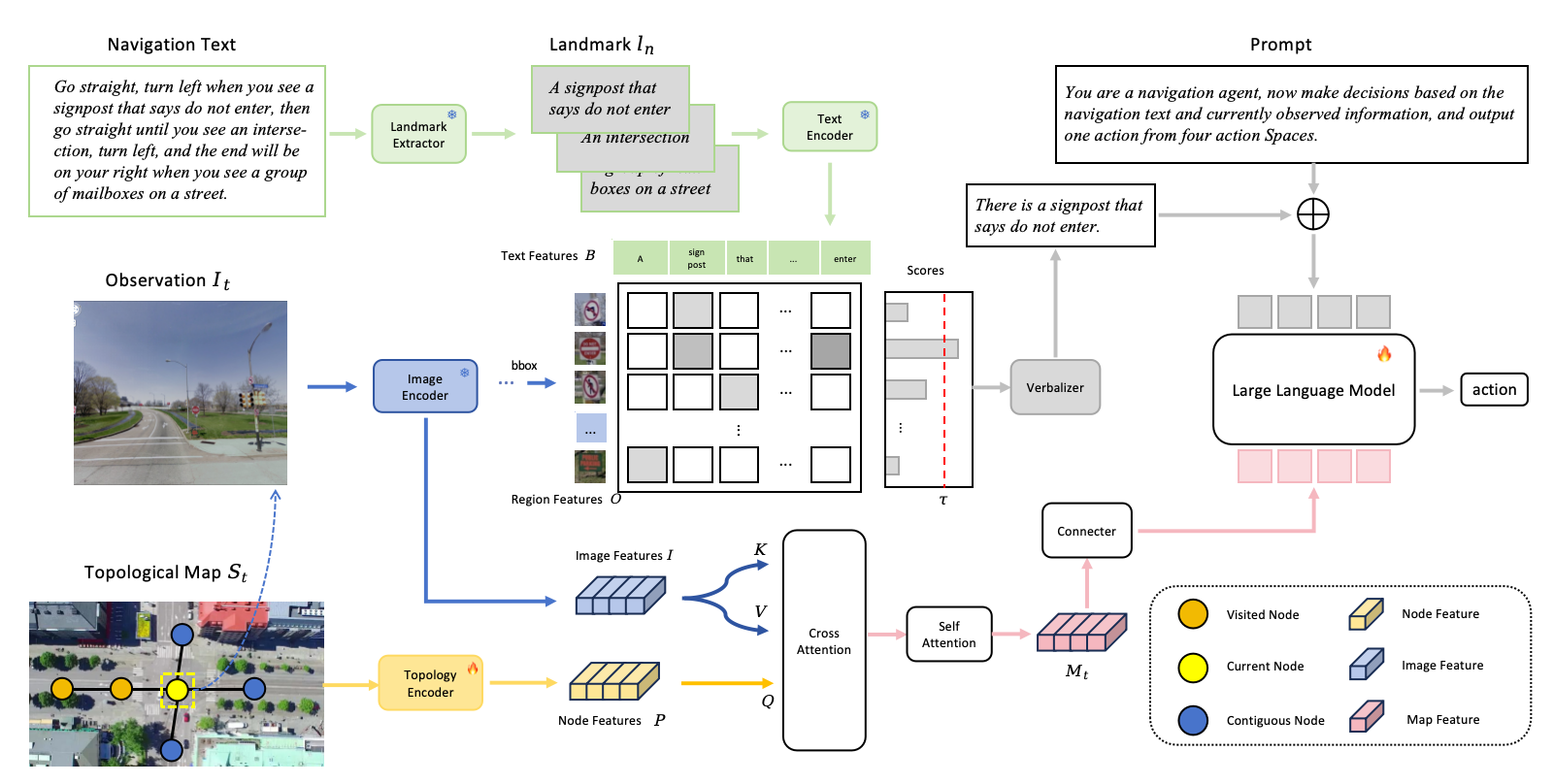}
	\caption{The overall pipeline. At step $t$, the region features $O$ extracted from the observation image $I_t$ and the text features $B$ of the landmark text extracted in the text extractor for landmark are computed to obtain the matching score, and then linguistically verbalized in the Verbalizer to obtain the landmark information $X$. The environmental topology map \( S \) is encoded by the topology map encoder to extract node features \( P \). The node features \( P \) and the current observation image features \( I \) are utilized to compute the global feature \( M_t \) through a cross-attention mechanism. Finally, the global feature \( M_t \) and the landmark information \( X \) are input into the LLM. After processing, the LLM generates action instructions.
 }
	\label{FIG:overview}
\end{figure*}
\renewcommand{\dblfloatpagefraction}{0.9}
\begin{figure}[h]
	\centering
		\includegraphics[scale=.4]{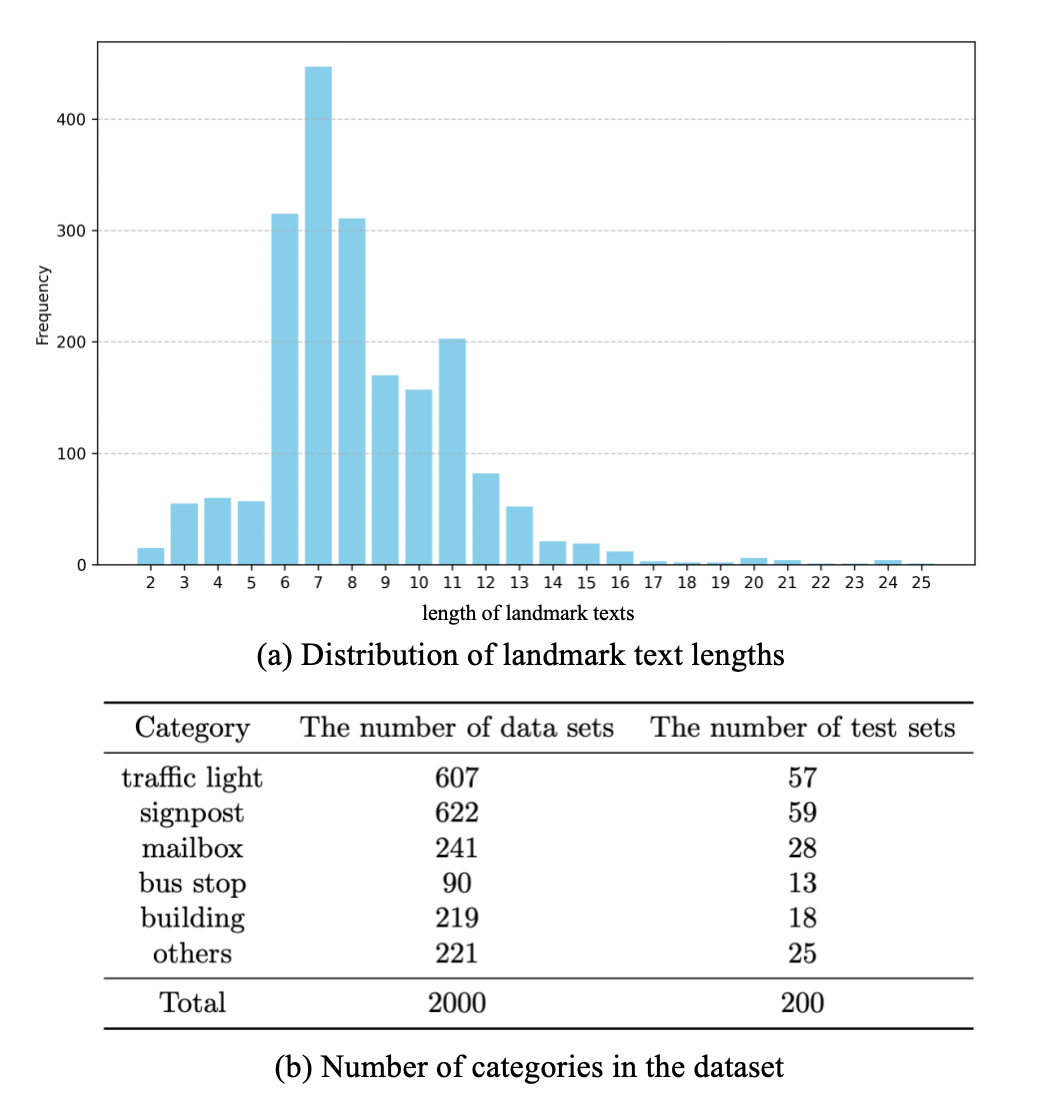}
	\caption{Figure (a) shows the distribution of landmark text lengths in the dataset, and Figure (b) shows the distribution of landmark types. 
 }
	\label{FIG:gen2}
\end{figure}

In this manner, we generate 2,000 one-to-one fine-grained landmark Image-Text pair data, which are
organized and synthesized into a cohesive dataset named NavAgent-Landmark2K. The distribution of the training set, testing set, and validation set in this dataset follows a 6:2:2 ratio. The average number of words for landmarks in the dataset is $8.24$, and the word length distribution graph is illustrated in Figure \ref{FIG:gen2} (a). The dataset contains six categories of landmarks, i.e., traffic lights, signposts, mailboxes, bus stops, buildings, and others, with the specific counts that are displayed in Figure \ref{FIG:gen2} (b).

\section{Methods}
\subsection{Task formulation}
In the VLN task, the starting node of the agent in the navigation environment is denoted as \( v_0 \). The initial orientation is represented as \( a_0 \), resulting in the initial state \( s_0 = \langle v_0, a_0 \rangle \). The model compute the next action \( a_1 \) based on the navigation instructions  \( T \) and the environmental panorama observed at the current node \( I_0 \). This is expressed as 
\begin{equation}
a_1 = F(s_0, I_0, T)
\end{equation}
where \( a_1 \) belongs to the action space within the navigation environment.
After calculating the resulting action \( a_1 \), the model updates the state of the agent based on the current state \( s_0 \) and the action \( a_1 \) to obtain the new state \( s_1 = \phi(a_1, s_0) \). This operation is repeated: when the agent moves to the new state \( s_t \), it then uses the new environmental information \( I_t \) to calculate the next action \( a_{t+1} \), expressed as:
\begin{equation}
a_{t+1} = F(s_t, I_t, T)
\end{equation}
followed by computing the next state \( s_{t+1} = \phi(a_{t+1}, s_t) \). This process continues until the agent computes the next action as STOP. The navigation task is deemed successful if the agent stops within the target node or a contiguous node of the target node.

\renewcommand{\dblfloatpagefraction}{0.9}
\begin{figure*}[t]
	\centering
		\includegraphics[width=\textwidth]{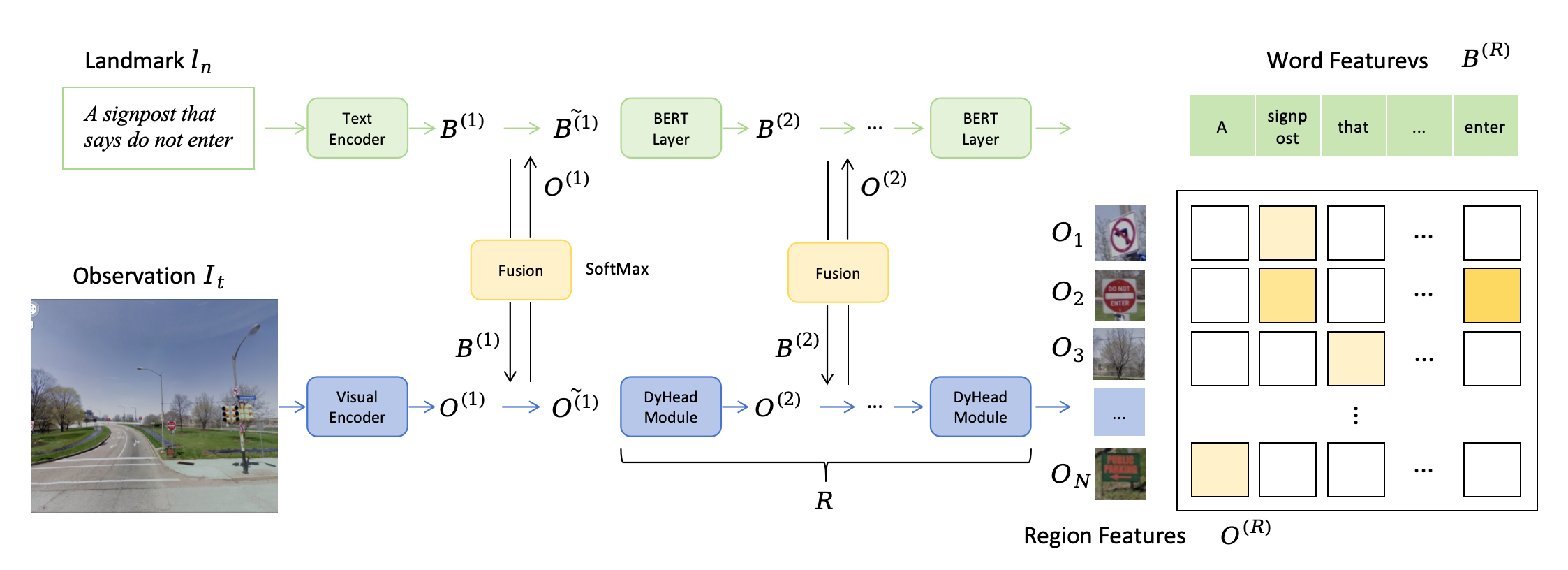}
	\caption{Schematic structure of visual recognizer for landmark. The figure illustrates the feature fusion process between the region feature \(O\) extracted from the observation image and the text feature \(B\) extracted from the landmark text. In the score matrix on the right side of the figure, darker colors indicate higher scores.
 }
	\label{FIG:glip}
\end{figure*}

\subsection{Overview}
UAV VLN tasks suffer from difficulties in encoding global environmental information for urban scenarios and low accuracy in recognizing complex fine-grained landmarks.
To address these issues and fill the research gap in UAV VLN based on multimodal large models in real-world scenarios, we propose the first embodied navigation model for urban drones driven by large Vision-Language model, which we name NavAgent. The architecture of the model is illustrated in Figure \ref{FIG:overview}.
NavAgent consists of four modules: the text extractor for landmark, the visual recognizer for landmark, the topological map encoder, and LLM. Initially, the text extractor for landmark retrieves the navigation instruction text $T$ to obtain the set of landmarks $L = \{l_1, l_2, \ldots, l_n\}$. Subsequently, the visual recognizer for landmark takes both the landmark set $L$ and the currently observed panoramic image $I_t$ as inputs, leveraging the landmark set to identify fine-grained targets within the panoramic image. the visual recognizer for landmark then extracts the information $X$ of the landmarks present at the node and inputs this information into the LLM. As the navigation progresses, the topological map is updated by integrating observations collected along the traversed paths. The topological map is then fed into the topological map encoder to extract topological map features $M_t$. These features are passed to the LLM through an adapter. Ultimately, the LLM receives the navigation instruction text $T$, environmental observations $I_t$, landmark information $X$, and topological map features $M_t$, and subsequently generates the action decision $a_{t+1}$.

\subsection{Text Extractor for Landmark}
There are multiple landmark phrases in the navigation instruction text \( T \) that prompt for turning. To determine whether the landmarks mentioned in \( T \) are visible in the current observation images, the first step is to extract these landmark phrases. We utilize a pre-trained LLM, which demonstrates remarkable emergence ability and performs well in zero-shot inference tasks, as our text extractor for landmark. This model will extract the landmark phrases from the text, resulting in \( L = \{ l_1, l_2, \ldots, l_n \} \). We design 10 cue prompts, each comprising three navigation instruction texts \( T \) and a corresponding set of manually extracted landmarks \( L \). During model training and inference, the parameters of the text extractor for landmark are frozen and executed before navigation begins. The extraction process is represented as:
\begin{equation}
L = \text{LLM}(T, \text{prompt})
\end{equation}

\subsection{Visual Recognizer for Landmark}
When the agent is at node \( t \), it observes the current environment to obtain a panoramic view \( I_t \). To verify that the landmarks \( l_i \) are visible at the current node, target recognition must be performed on the panorama. We organize the panorama \( I_t \) into three images based on the left, front, and right sides, each with a 60-degree viewing angle, denoted as \( I_t^1 \), \( I_t^2 \), and \( I_t^3 \).
To address the challenges of recognizing fine-grained landmarks in observation images and complex landmarks in navigation texts, we design a visual recognizer for landmark. This recognizer is based on the GLIP and has been fine-tuned using the NavAgent-Landmark2K dataset. The structure diagram is illustrated in Figure \ref{FIG:glip}. It sequentially matches the images from the three viewpoints with the landmark  \( l_n \). For example, in the front view, \( N \) bounding boxes are generated in the image, and the image features \( O_i^{(1)} \) are extracted by the image encoder. Then, the text features \( B^{(1)} \) are extracted by the text encoder, expressed as:
\begin{equation}
O_i^{(1)}=Enc_I (Img),i \in {0,1,...,N-1}
\end{equation}
\begin{equation}
B^{(1)}=Enc_T (l_n)
\end{equation}

To enable the model to learn accurate phrase-level matching capabilities, a deep fusion of visual and text features is required. Specifically, the two features are first interacted separately using cross-attention to obtain:
\begin{equation}
\tilde{O_i^{(1)}} =softmax((O_i^{(1)} W_q (B^{(1)} W_k )^T)/\sqrt{d}) B^{(1)} W_v
\end{equation}
\begin{equation}
\tilde{B^{(1)}} =softmax((B^{(1)} W_q (O_i^{(1)} W_k )^T)/\sqrt{d}) O_i^{(1)} W_v
\end{equation}
The new visual and text features are then input into DyHead\cite{9577765} and BERT\cite{Devlin2019BERTPO}, respectively, to fuse with the original features, resulting in:
\begin{equation}
O_i^{(2)}=DyHEadModule(O_i^{(1)}+\tilde{O_i^{(1)}} )
\end{equation}
\begin{equation}
B^{(2)}=BERTLayer(B^{(1)}+\tilde{B^{(1)}} )
\end{equation}
This process is repeated \( R \) times to obtain \( O^{(R)} \) and \( B^{(R)} \). Finally, the similarity is calculated using the visual and text features that have undergone multiple cross-fusions to obtain:
\begin{equation}
Score_t [l_n ]=max_{0 \leq i<N-1}{O_i^{(R)} {B^{(R)}}^T }
\end{equation}

Additionally, the visual recognizer for landmark includes a component called the verbalizer, which converts the results of environmental observations into textual form for input into the LLM. This process is based on the landmark recognition scores output from the GLIP. For example, when the score \( \text{Score}_t[l_n] \) exceeds a set threshold $\tau$, the verbalizer will output the text “There is [$l_n$] on your [$d_i$]”.

\subsection{Topology Map Encoder}
\renewcommand{\dblfloatpagefraction}{0.9}
\begin{figure*}[t]
	\centering
		\includegraphics[width=\textwidth]{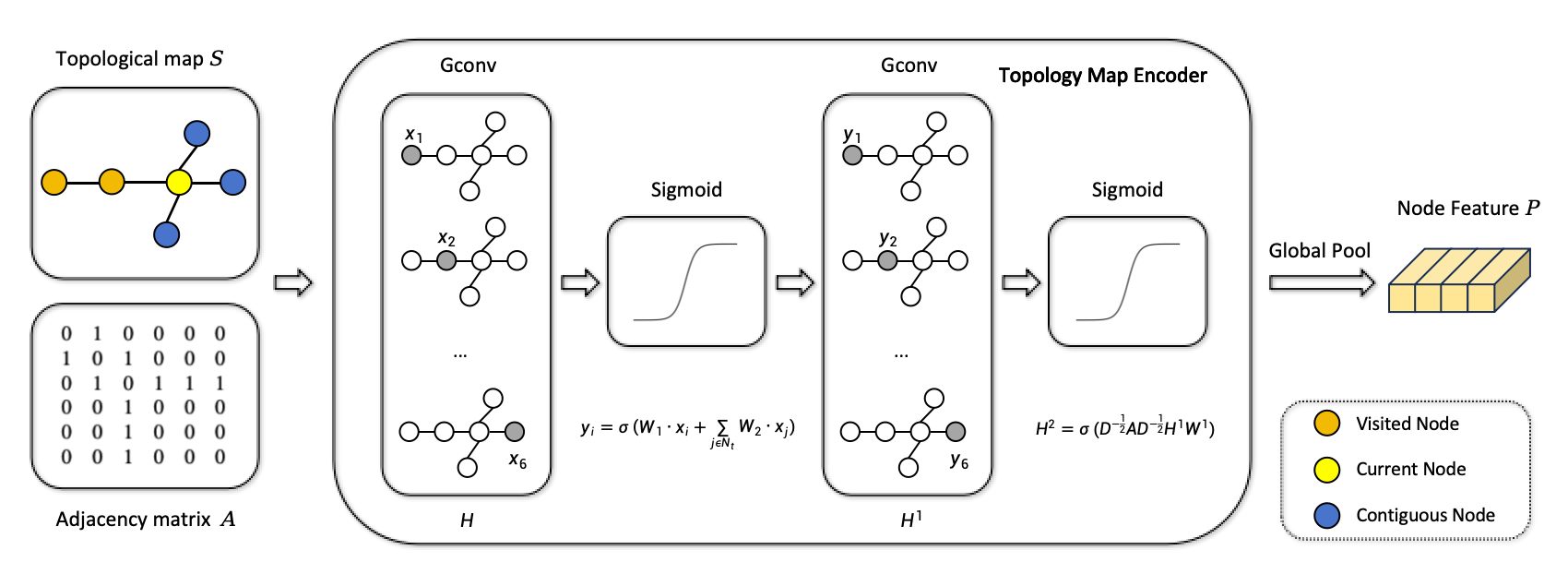}
	\caption{Schematic structure of Topology Encoder. The scene topology map \( S \) is processed through two graph convolution layers, resulting in the node feature matrix \( H^2 \). Subsequently, global pooling is applied to derive the global node features \( P \).
 }
	\label{FIG:topology}
\end{figure*}
To enhance the ability of agent to understand spatial relationships during navigation and improve long-term planning capacity, we construct a dynamically growing scene topology map that contains the observed environmental locations, denoted as \( S \).
The topology map begins traversing the visited and contiguous nodes starting from the initial node and is abstracted into a graphical representation, denoted at step \( t \) as \( S_t = \langle N_t, E_t \rangle \), where \( N_t \) is the set of nodes contained in the topology map, and \( E_t \) is the set of edges connecting two nodes within the map. We categorize the nodes \( n_i \in N_t \) into three categories: visited nodes, current node, and contiguous nodes.

After obtaining the scene topology \( S_t \), each node and its spatial relationships are encoded using a Topology Map Encoder. Specifically, we utilize a GCN for feature aggregation, updating each node with information from all nodes in the topology map. This allows us to refine the features of each node, represented as:
\begin{equation}
 y_i= \sigma(W_1·x_i + \sum_{j \in N_t}W_2·x_j)
\end{equation}where \( x_i \) denotes the initial features of node \( n_i \), and \( W_1 \) and \( W_2 \) are the learnable parameters.

At this point, the node feature matrix is represented as \( H^1 = [y_1, y_2, \ldots, y_N] \). To enable nodes to gather information from their distant neighbors and capture global features, we stack two GCNs to aggregate information progressively. This process is represented by the computation
\begin{equation}
H^2= \sigma(D^{-_2^1} AD^{-_2^1 } H^1 W^1 )
\end{equation}
where \( D^{-_2^1} AD^{-_2^1 } \) is the normalized adjacency matrix of the topological map. In the third layer, the node features are aggregated into global node features using global pooling, denoted as:
\begin{equation}
P = \text{pool}(H^2)
\end{equation}

The structure of the Topology Map Encoder is schematically shown in Figure \ref{FIG:topology}. To enable the model to concentrate on the information of the current node, we input the panorama of the environment observed by the current node into the image encoder to extract the image features \( O_t \). Subsequently, the node features \( P \) and the image features \( O_t \) are fed into the Cross Attention layer to facilitate cross-modal interaction, yielding the topological map features \( M_t \) that contain the environmental information. This results in the topological map features being represented as: 
\begin{equation}
M_t=softmax((PW_q (O_t W_k )^T)/\sqrt{d}) O_t W_v 
\end{equation}

\subsection{LLM}
After extracting the global topological map features \( M_t \) in the scene topology map \( S \), we encounter modal gaps that prevent \( M_t \) from being directly input into the LLM. Therefore, a connector module is required to achieve modal alignment. We employ a Multi-Layer Perceptron (MLP) as a learnable projection network, through which the global topological map features \( M_t \) are mapped to obtain \( M_t^\prime = projection(M_t) \).
In this manner, the LLM can process the global environmental topological map features that contain relevant environmental information and combine them with the local landmark recognition results and navigation text to generate navigation decisions. 
The process of inference using LLM is shown in Algorithm \ref{alg:infer}.

 \begin{algorithm}[!h]
    \caption{The inference process of NavAgent}
    \label{alg:infer}
    \renewcommand{\algorithmicrequire}{\textbf{Input:}}
    \renewcommand{\algorithmicensure}{\textbf{Output:}}
    \begin{algorithmic}[1]
        \REQUIRE for Node $t$, Navigation Text $T$, Observation Image $I_t$, Topology map $S_t$.  
        \ENSURE Action   
        
        \STATE  $t \leftarrow 1$, Action $\leftarrow$ None
        \WHILE{Action $\ne$ STOP}
            \STATE $E_l$ = text extractor for landmark
            \STATE $E_T$ = Text Encoder
            \STATE $E_I$ = Image Encoder
            \STATE $E_G$ = Topology Encoder
            \STATE landmarks $l_n$ $\leftarrow$ $E_l(T)$
            \STATE $Feature_{l_n} \leftarrow E_T(l_n)$
            \STATE Image Feature $I$, Region Feature $O$ $\leftarrow E_I(I_t)$
            \STATE Node Feature $P$ $\leftarrow E_G(S_t)$
            \STATE $X$ $\leftarrow O \cdot Feature_{l_n}$
            \STATE $M_t \leftarrow Cross\_Attention(P,I)$
            \STATE Action  $\leftarrow LLM(X,M_t)$
            \STATE  $t \leftarrow t+1$
        \ENDWHILE
    \end{algorithmic}
\end{algorithm}

\subsection{Loss Function}
During the training process, to enable the agent to learn how to synthesize global and local information for decision-making in navigation, we utilize the loss function of the LLM, denoted as \( \text{Loss}_{\text{llm}} \). Based on the topological map \( S \) of the scene generated by the agent at time \( t \) with the ground truth value \( C \), we calculate the topological map loss, \( \text{Loss}_T \). The topological map needs to be transformed into the adjacency matrix \( A \) during the calculation process. The total loss function and the topological map loss are defined as follows:
\begin{equation}
Loss_T = {\lVert{A_S - A_C}\rVert}^2
\end{equation}
\begin{equation}
Loss = \lambda_1 Loss_T + \lambda_2 Loss_{llm}
\end{equation}

\section{Experiments}
\subsection{Experimental Setup}
\textbf{Implementation details}. This experiment is divided into two phases. In the first phase, we fine-tune the GLIP using the NavAgent-Landmark2K dataset and evaluate the fine-tuned model on a fine-grained landmark recognition task. The initial weights of the GLIP are based on the glip\_tiny\_model\_o365\_goldg\_cc\_sbu version.
In the second phase, we train NavAgent on the Touchdown and Map2seq datasets and evaluate its performance in unseen scenarios. We utilize GPT-4 as the text extractor for landmark, the GLIP trained in the first phase as the visual recognizer for landmark, and the LLaMa2-13b model as the LLM for decision-making. The output threshold in the verbalizer module is set to 0.8, and the $\tau$ value is also set to 0.8. In the loss function calculation, we set $\lambda_1$ and $\lambda_2$ to 0.5 to balance the two losses. In the first phase, we fine-tuned the GLIP for 25 epochs with a learning rate of 0.0001, where each epoch took 3 hours on an NVIDIA 3090 Ti GPU. In the second phase, we trained the model for 20 epochs using LoRA, with lora\_r set to 8 and a learning rate of 0.0003. Each epoch in this phase took 1 hour on an 8-card NVIDIA A800 GPU.

\textbf{Evaluation metrics}. Three metrics are selected for this experiment to evaluate the performance of the VLN task: Task Completion Rate (TC), Shortest Path Distance (SPD), and Key Point Accuracy (KPA). 

Task Completion Rate (TC) refers to the proportion of instances where the agent stops within one contiguous node of the target location. 

Shortest Path Distance (SPD) measures the length of the shortest path between the stopping position of the agent and the target position\cite{Chen2018TOUCHDOWNNL}.

Key Point Accuracy (KPA) focuses on the decision-making ability of the agent at key points during the navigation process. It is calculated as the rate of correct decisions made at these key points, which include initial nodes, nodes with landmarks, and the target node. 

The evaluation metrics are formulated as follows:
\begin{equation}
TC = Num_{success}/Num_{all}
\end{equation}
\begin{equation}
SPD = min_{distance}(Loc_{goal}-Loc_{stop})
\end{equation}
\begin{equation}
KPA = Num_{success\ in\ Keypoint}/Num_{all\ in\ Keypoint}
\end{equation}

\textbf{Baselines}. We select several representative classes of models as baselines and fine-tune some models for the UAV embodied navigation task. 

(1) Miniature Model: ORAR employs a sequence-to-sequence architecture\cite{schumann2022analyzinggeneralizationvisionlanguage}, where an LSTM serves as the encoder to read the navigation instruction text, while another LSTM functions as the multilayer decoder, receiving the image feature vector of the current panoramic view to enhance each action decoding step.

(2) Large Language Model (LLM): The LLM cannot directly receive image features and environmental information. Therefore, we reference VELMA's linguistic workflow to convert environmental data into text using a verbalizer, and input to the LLM for inference. We adapt the GPT-3, GPT-4, and Vicuna models, providing two contextual examples without fine-tuning. One of these, VELMA, is the state-of-the-art model for urban VLN agents, utilizing the verbalization of trajectories and visual environment observations as contextual cues for subsequent actions.

(3) Large Vision-Language Model (VLM): The VLM can process both image and text inputs. At each node, we input the forward observation images along with the overall navigation text into the VLM. This process loads the pre-trained model weights directly, without fine-tuning. For our study, we select the GPT-4-vision, GPT-4o, BLIP2, and LLaVA models.
\begin{table*}[hbt]
\centering
\caption{Evaluation results on the Touchdown and Map2seq datasets.}
\label{result}
\begin{tabular}{c|ccc|ccc|ccc|ccc}
\toprule
\phantom{} & \multicolumn{6}{c|}{\centering Touchdown} & \multicolumn{6}{c}{\centering Map2seq} \\
\midrule
\phantom{} & \multicolumn{3}{c|}{Development Set} & \multicolumn{3}{c|}{Test Set} & \multicolumn{3}{c|}{Development Set} & \multicolumn{3}{c}{Test Set} \\
\midrule
\textbf{Models$\downarrow$} & SPD$\downarrow$ & KPA$\uparrow$ & TC$\uparrow$ & SPD$\downarrow$ & KPA$\uparrow$ & TC$\uparrow$ & SPD$\downarrow$ & KPA$\uparrow$ & TC$\uparrow$ & SPD$\downarrow$ & KPA$\uparrow$ & TC$\uparrow$ \\
\midrule
&\multicolumn{11}{c}{\textbf{ $ \textnormal{Miniature Model}$}}& \\
\midrule
ORAR & 20.0 & - & 15.4 & 20.8 & - & 14.9 & 11.9 & - & 27.6 & 13.0 & - & 30.3 \\
\midrule
&\multicolumn{11}{c}{\textbf{ $ \textnormal{Large Language Model}$}}& \\
\midrule
GPT-3 & 22.2 & 49.1 & 6.8 & - & - & - & 19.1 & 58.1 & 9.2 & - & - & - \\
GPT-4 & 21.8 & 56.1 & 10.0 & - & - & - & 12.8 & 70.0 & 23.1 & - & - & - \\
Vicuna & 22.9 & 51.4 & 7.5 & - & - & - & 17.4 & 60.8 & 11.6 & - & - & - \\
VELMA & 15.5 & 63.6 & 26.0 & 16.0 & 62.8 & 26.4 & 8.3 & 79.5 & 45.3 & 8.3 & 79.6 & 47.5 \\
\midrule
&\multicolumn{11}{c}{\textbf{ $ \textnormal{Large Vision-Language Model}$}}& \\
\midrule
GPT-4-vision & 21.5 & 48.6 & 8.5 & 20.9 & 49.0 & 7.9 & 13.3 & 69.1 & 21.8 & 12.8 & 68.4 & 22.0 \\
GPT-4o & 20.4 & 49.2 & 8.7 & 20.5 & 51.4 & 9.3 & 12.5 & 71.4 & 25.3 & 12.2 & 71.5 & 25.1 \\
BLIP2-flan-t5-xxl & 23.4 & 45.6 & 5.6 & 25.4 & 43.8 & 4.9 & 18.1 & 58.9 & 12.1 & 17.5 & 60.1 & 13.4 \\
BLIP2-opt-6.7b & 23.9 & 44.5 & 5.3 & 24.8 & 44.9 & 5.6 & 20.0 & 58.2 & 11.8 & 19.2 & 58.9 & 12.5 \\
LLaVA & 22.8 & 45.1 & 6.2 & 21.1 & 44.7 & 6.1 & 17.3 & 63.7 & 13.5 & 17.8 & 63.9 & 13.3 \\
\midrule
NavAgent & \textbf{14.1} & \textbf{65.2} & \textbf{27.2} & \textbf{14.9} & \textbf{63.4} & \textbf{27.0} & \textbf{7.8} & \textbf{80.5} & \textbf{46.4} & \textbf{8.0} & \textbf{81.4} & \textbf{47.9} \\
\bottomrule
\end{tabular}
\end{table*}

\subsection{Performances of Text Extractor for Landmark}
To ensure that the selected text extractor for landmark performs optimally in extracting landmark phrases, we select several pre-trained LLMs, including GPT-3, GPT-4, and LLaMa. The test data are sourced from the Touchdown and Map2seq datasets, with 50 navigational texts selected from each. To enhance the accuracy of the evaluation, we manually label the landmark phrases in these texts, resulting in an average of three landmark phrases per text. The phrase scores extracted by each bigram model are presented in Table \ref{f1}. All models demonstrate relatively excellent performance, even though they are not specifically trained. In particular, GPT-4 exhibits outstanding performance in landmark phrase extraction scores, which leads us to select it as the base model for the text extractor for landmark.

\begin{table}[h]
\centering
\caption{Results of different LLMs in the task of landmark phrase extraction.}
\label{f1}
\begin{tabular}{c|ccc|ccc}
\toprule
\phantom{} & \multicolumn{3}{c|}{\centering Touchdown} & \multicolumn{3}{c}{\centering Map2seq} \\
\midrule
\phantom{} & Precision & Recall & F1 & Precision & Recall & F1 \\
\midrule
GPT-3 & 97.5 & 94.9 & 96.2 & 98.6 & 97.3 & 97.9\\
GPT-4 & \textbf{98.4} & \textbf{98.2} & \textbf{98.3} & \textbf{99.6} & \textbf{99.7} & \textbf{99.6}\\
LLaMa2-13b & 98.0 & 96.1 & 97.1 & 98.7 & 97.4 & 98.1\\
\bottomrule
\end{tabular}
\end{table}

\subsection{Performances of Visual Recognizer for Landmark}
\begin{figure}[h]
	\centering
		\includegraphics[scale=.2]{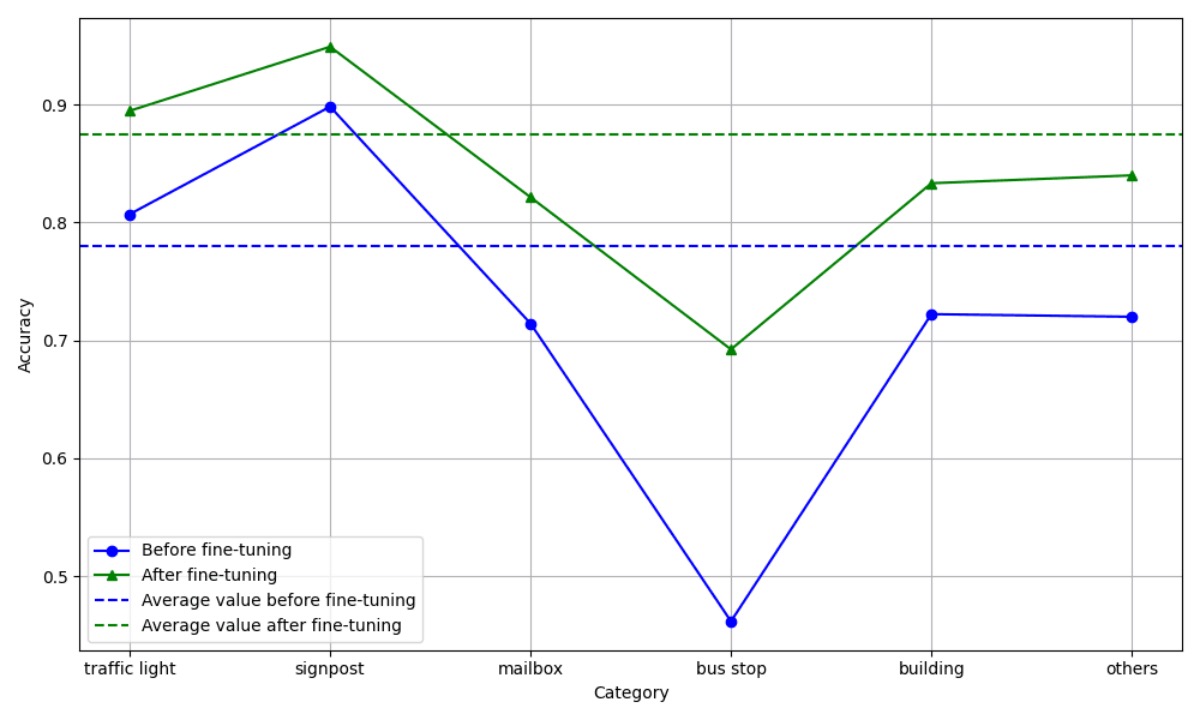}
	\caption{Fine-grained landmark recognition accuracy before and after fine-tuning. 
 }
	\label{FIG:glip_result}
\end{figure}

The accuracy curves of the GLIP before and after fine-tuning on the NavAgent-Landmark2K validation set are presented in Figure \ref{FIG:glip_result}. The experimental results indicate that the fine-tuned GLIP, trained using our NavAgent-Landmark2K dataset, demonstrates exceptional performance in the fine-grained landmark recognition task. It can accurately identify landmarks that occupy a relatively small percentage of the complex scene, thereby facilitating the ability of visual recognizer for landmark to convert the observed image information into landmark recognition data. After fine-tuning, the overall recognition accuracy improved by $9.5\%$. Additionally, we calculate the recognition accuracies across different landmark categories, and the results indicate significant improvements for each category. In particular, the recognition accuracy for the bus stop category increased by $23.1\%$.

\renewcommand{\dblfloatpagefraction}{0.9}

\subsection{Quantitative Results}
\renewcommand{\dblfloatpagefraction}{0.98}
\begin{figure*}[t]
	\centering
 
		\includegraphics[width=0.9 \textwidth]{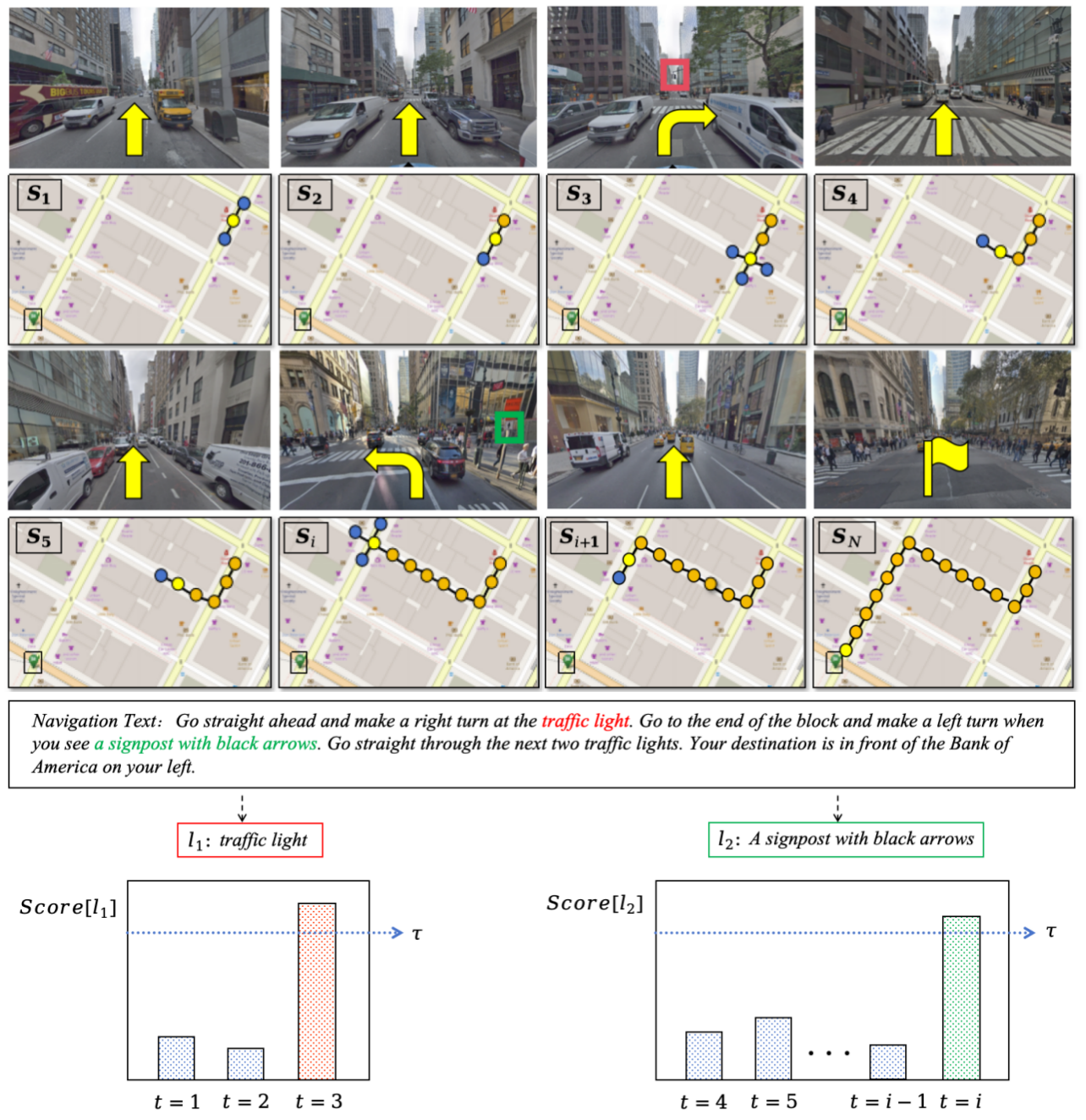}
	\caption{Visualization results of navigation examples of NavAgent. The top section of the figure displays the observed images and scene topology during the NavAgent navigation. The center section presents the navigation text, while the bottom section illustrates the scores for the landmark categories.
 }
	\label{FIG:case}
\end{figure*}
\textbf{Comparison With the Baseline}. In this section, we compare the NavAgent model with several representative previous models, as well as State-of-the-Art (SOTA) models. As shown in Table \ref{result}, the results indicate that NavAgent exhibits outstanding performance on both the Touchdown and Map2seq datasets. Specifically, on the Touchdown dataset, NavAgent improves the task completion rate by $4.6\%$ and $2.2\%$ compared to VELMA on the development and test sets, respectively. Furthermore, on the Map2seq dataset, NavAgent achieves improvements of $2.4\%$ and $0.8\%$ over VELMA on the development and test sets, respectively. Additionally, NavAgent outperforms other baseline models in terms of the SPD and KPA metrics, further demonstrating the effectiveness of our model. It is also worth noting that the large Vision-Language Model relies solely on the observation image of the current node and navigation text, without taking into account the spatial relationships and historical information between different nodes. Consequently, it performs poorly in zero-shot scenarios.
\begin{table}[h]
\centering
\caption{Results of Ablation experiments.}
\label{ablation}
\begin{tabular}{c|ccc|ccc}
\toprule
\phantom{} & \multicolumn{3}{c|}{\centering Touchdown} & \multicolumn{3}{c}{\centering Map2seq} \\
\midrule
\phantom{} & SPD$\downarrow$ & KPA$\uparrow$ & TC$\uparrow$ & SPD$\downarrow$ & KPA$\uparrow$ & TC$\uparrow$ \\
\midrule
w/o GLIP & 16.7 & 62.9 & 24.3 & 10.5 & 76.9 & 43.5\\
w/o Map & 15.3 & 64.1 & 26.4 & 8.2 & 79.7 & 45.0\\
NavAgent & \textbf{14.1} & \textbf{65.2} & \textbf{27.8} & \textbf{7.8} & \textbf{80.5} & \textbf{46.4}\\
\bottomrule
\end{tabular}
\end{table}

 



\textbf{Ablation Study}. To verify the effectiveness of each module, we conduct ablation experiments on NavAgent, with the results presented in Table \ref{ablation}. 
First, we remove the visual recognizer for landmark and input only the topological map features, which encompass environmental information and current observations, into the LLM. The performance degradation of the model is evident in the w/o GLIP results, as the model lacks the ability to recognize fine-grained landmarks during the decision-making process, making it challenging to ascertain whether turning is required at the current node. 
Next, we eliminate the topological map encoder module and solely utilize the visual recognizer for landmark to identify the current node landmarks, which are then fed into the LLM via the verbalizer. The model’s performance declines, as evidenced by the w/o Map results. This decline is attributed to the model’s inability to comprehend the spatial relationships between different nodes and the topology, adversely affecting its long-term path planning and adjustment capabilities.


\renewcommand{\dblfloatpagefraction}{0.98}
\begin{figure*}[t]
	\centering
		\includegraphics[width=\textwidth]{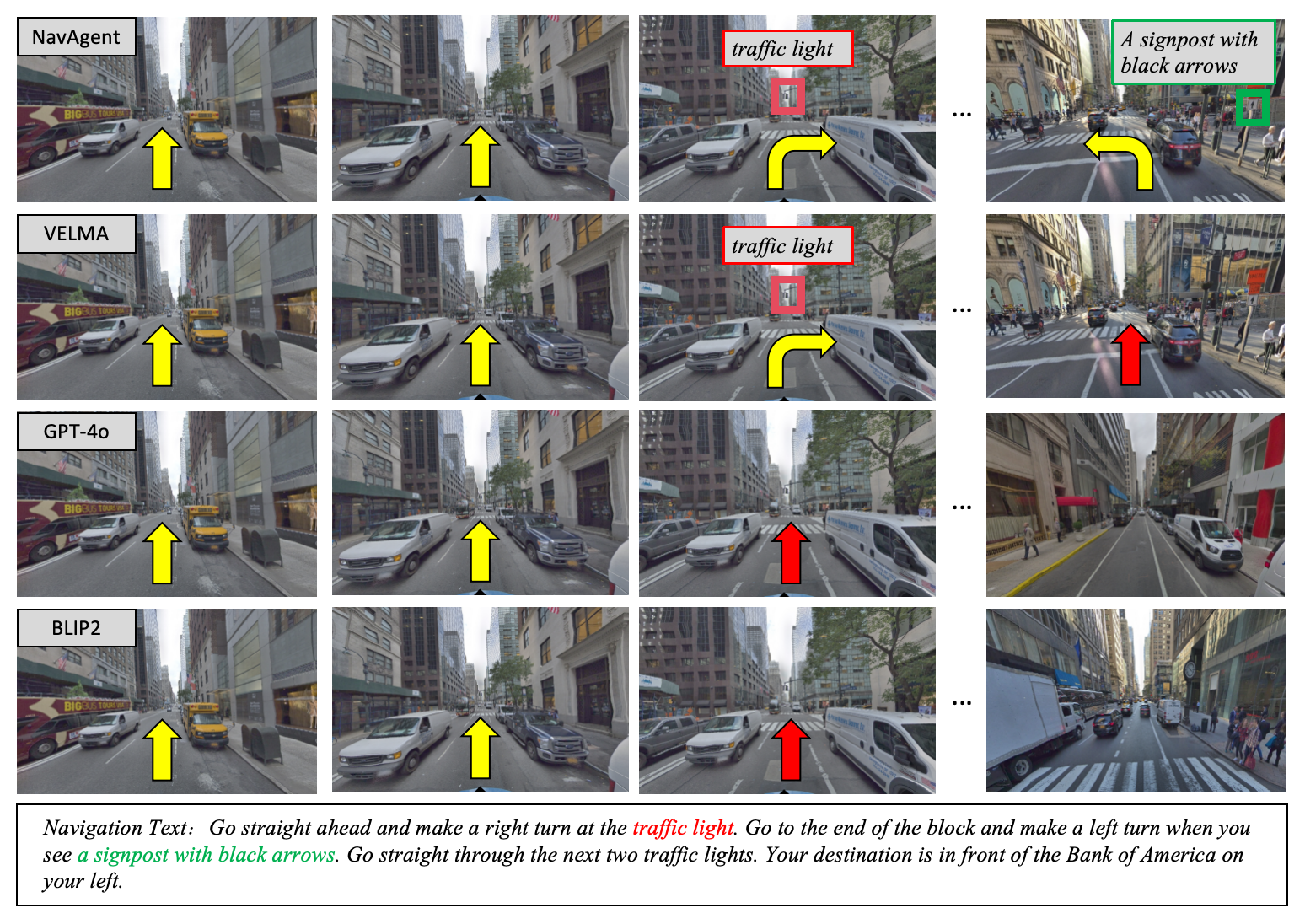}
	\caption{Visualization results of navigation examples of NavAgent and other baselines. Yellow arrows indicate correct decisions, while red arrows signify incorrect decisions.
 }
	\label{FIG:lizi}
\end{figure*}

\begin{figure}[h]
	\centering
		\includegraphics[scale=.55]{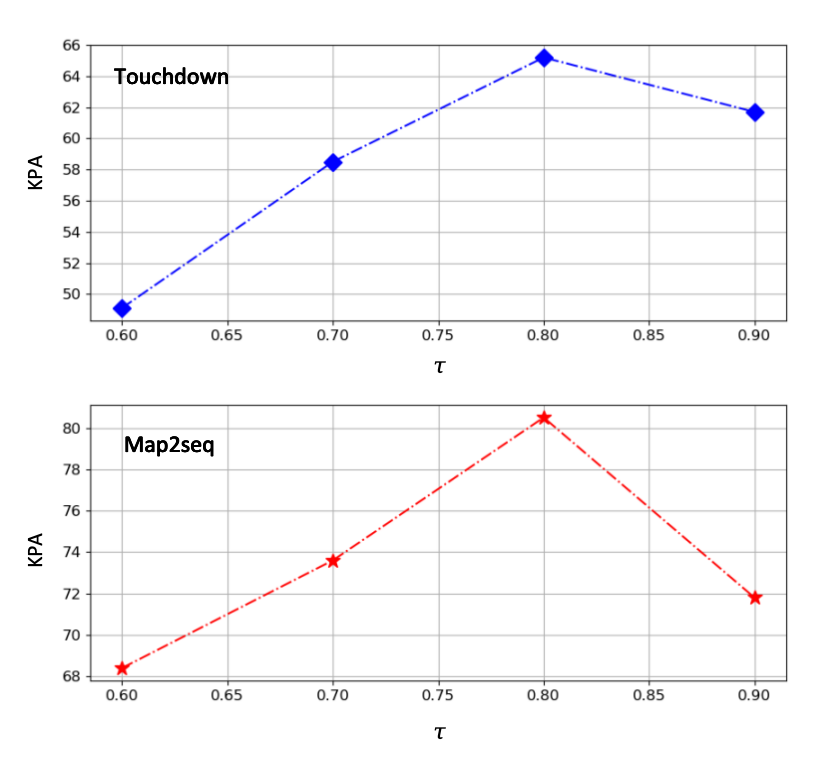}
	\caption{Effect of different $\tau$ on KPA on the Touchdown and Map2seq datasets. The values of $\tau$ are 0.6, 0.7, 0.8, and 0.9, respectively.
 }
	\label{FIG:ablation}
\end{figure}

Additionally, we conduct a thorough investigation of the output threshold in the visual recognizer for landmark, denoted as $\tau$, and its impact on the navigational performance of the agent. Specifically, we set $\tau$ values to 0.6, 0.7, 0.8, and 0.9 to evaluate keypoint accuracy on both the Touchdown and Map2seq datasets. The results are shown in Figure \ref{FIG:ablation}. When $\tau$ is set to a lower value, the visual recognizer for landmark tends to misrecognize other objects as landmarks extracted from the navigation text, which causes the agent to turn too early. Conversely, when $\tau$ is set to a higher value, the visual recognizer for landmark is prone to overlook fine-grained targets appearing in the observation image, resulting in the agent missing the node it should turn.
\renewcommand{\dblfloatpagefraction}{0.95}

\subsection{Qualitative Results}
To visually demonstrate the effectiveness of our approach in VLN, we present the results of a navigation example visualization of NavAgent in Figure \ref{FIG:lizi}. The NavAgent matches landmarks at each node against those extracted by the current text extractor for landmark \( l_i \). As shown in the table of the Figure \ref{FIG:lizi}, the visual recognizer for landmark can accurately identify fine-grained landmarks in the observed image. 
When \( t = 3 \), the landmark to be matched is \( l_1\):“\textit{traffic light}", and the matching score calculated by the visual recognizer for landmark is denoted as \( \text{Score}_3[l_1]=0.91 \textgreater \tau \). The output message \( X_3 \) is “\textit{There is a traffic light}". The LLM combines message \( X_3 \) with the environment's topology at that moment \( S_3 \), calculating that the next action is to turn right. 
When \( t = i \), the landmark to be matched is \( l_2 \):“\textit{A signpost with black arrows}", and the matching score calculated by the visual recognizer for landmark is denoted as \( \text{Score}_i[l_2]=0.83 \textgreater \tau \). The output message \( X_i \) is “\textit{There is a signpost with black arrows}". The LLM combines message \( X_i \) with the environment's topology at that moment \( S_i \), calculating that the next action is to turn left.

To provide a more illustrative comparison between the NavAgent and other baselines, we have selected a representative example for in-depth analysis. As shown in Figure \ref{FIG:case}, when the agent reaches the first critical node in the navigation process, the landmark $l_1$ to be matched is “\textit{traffic light}”. This landmark phrase lacks modifiers, making it easier to recognize. At this stage, both the NavAgent and the VELMA model successfully recognize the landmark from the observed image and make correct decisions accordingly. However, large Vision-Language Models such as GPT-4o and BLIP2 fail to make turning decisions because they make predictions based solely on the observation images and navigation text, without specific training for this task. 
At the second critical node of the navigation process, the landmark $l_2$ to be matched is “\textit{A landmark with black arrows}”. This landmark phrase is complex and occupies a small percentage of the image, which causes VELMA to fail to recognize it correctly, resulting in a poor decision. This example demonstrates the differences between the various models in landmark recognition ability and decision-making processes, further validating the advantages of the NavAgent.



\section{Conclusion}
In this work, we propose NavAgent, the first urban UAV embodied navigation model driven by a large Vision-Language Model. It utilizes a visual recognizer for landmark to extract local information from landmarks in observed images, a topological map encoder to incorporate global environmental information alongside current visual information, and an LLM to synthesize multi-scale information effectively. In addition, we construct NavAgent-Landmark2K, the first fine-grained landmark dataset for real urban street scenes. Finally, we evaluated NavAgent both quantitatively and qualitatively on the Touchdown and Map2seq datasets. The results demonstrate superior performance compared to current state-of-the-art methods, thereby confirming the effectiveness of our approach in UAV VLN tasks.

In our future work, we plan to enhance the proposed method to improve the navigation capabilities of the embodied UAV agent in real-world scenarios. We aim to increase the stability of navigation under practical challenges, such as complex road conditions and pedestrian obstacles. Additionally, we intend to extend the functionality of NavAgent to support real-time human updates and adjustments during navigation.

\bibliography{ref}
\bibliographystyle{IEEEtran}
  \begin{IEEEbiography}[{\includegraphics[width=1in,height=1.25in,clip,keepaspectratio]{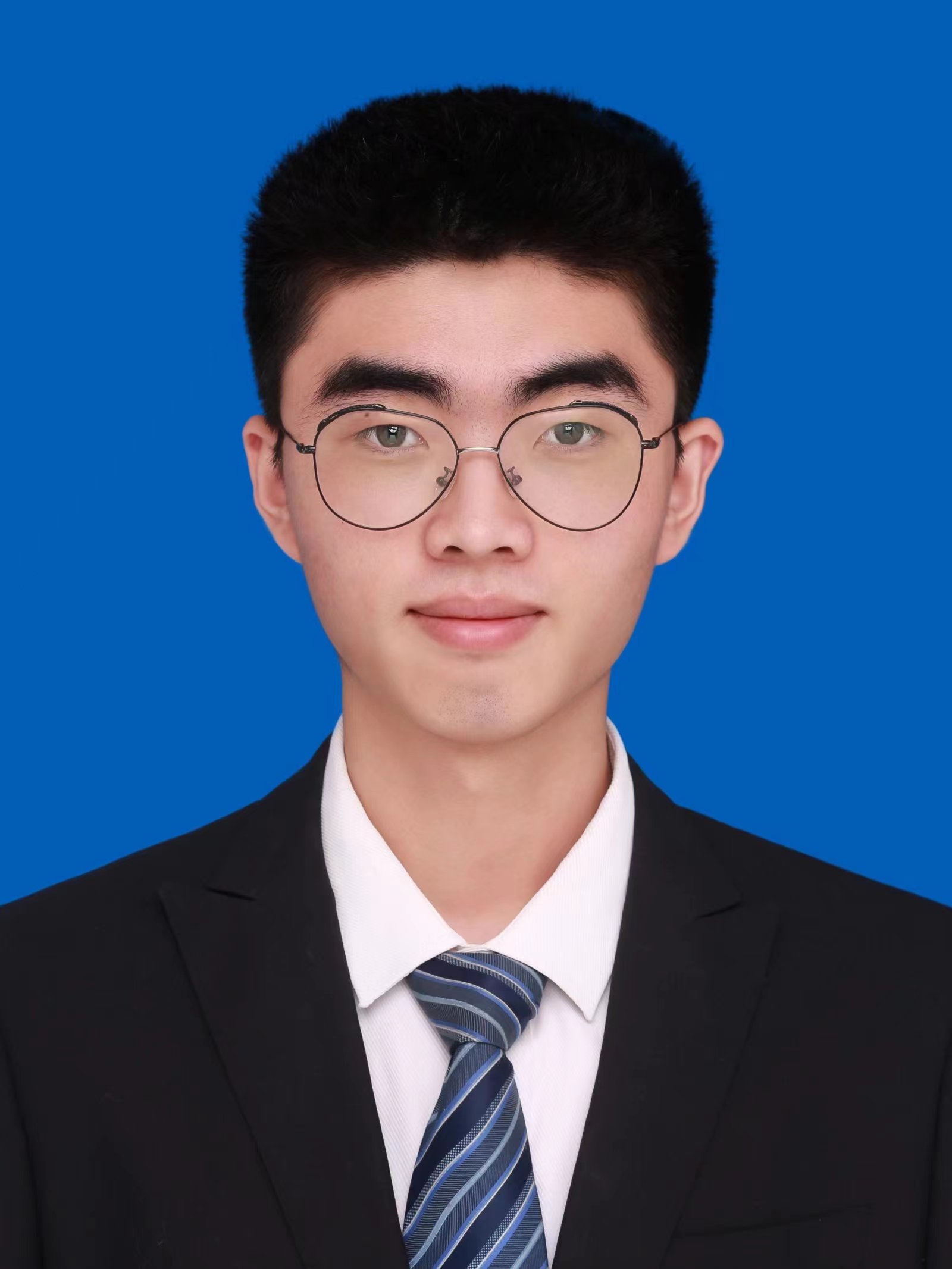}}]{Youzhi Liu}
        received the B.Sc. degree from Hunan University, changsha, China, in 2022. He is currently a Ph.D student with the Aerospace Information Research Institute, Chinese Academy of Sciences. 

        His research interests include embodied intelligence, and Vision-and-Language Navigation. 
	\end{IEEEbiography}
 \vspace{-1cm}
 
\begin{IEEEbiography}[{\includegraphics[width=1in,height=1.25in,clip,keepaspectratio]
{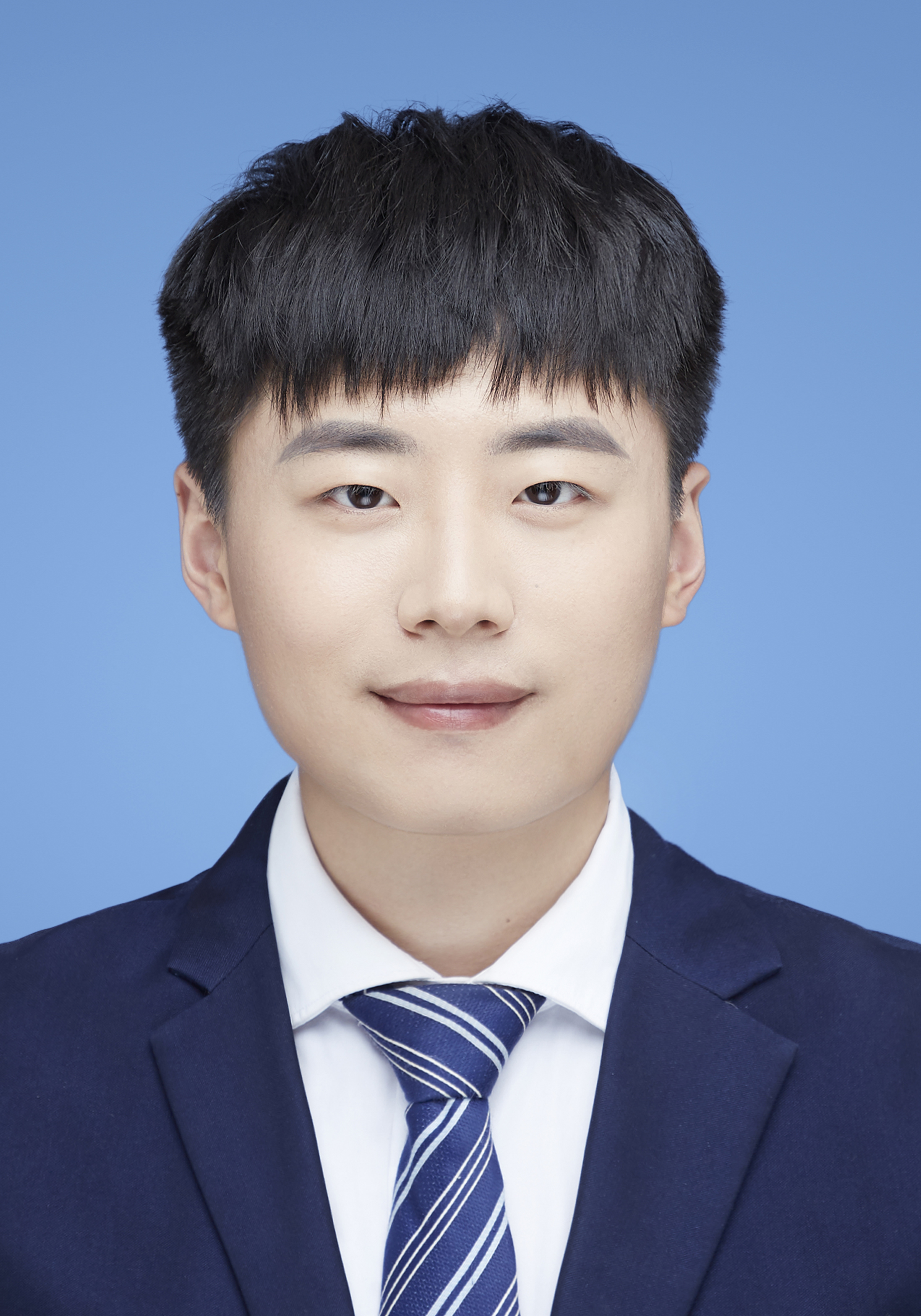}}]{Fanglong Yao}
        received the B.Sc. degree from Inner Mongolia University, Hohhot, China, in 2017, and the Ph.D. degree from the Aerospace Information Research Institute, Chinese Academy of Sciences, Beijing, China, in 2022. He is currently a Post-Doctoral Researcher and assistant professor with the Aerospace Information Research Institute, Chinese Academy of Sciences. 

        His research interests include cognitive intelligence, embodied intelligence, and swarm intelligence, concentrating on multi-agent learning, multimodal fusion, 3D scene understanding and spatiotemporal data analysis. 
	\end{IEEEbiography}
 \vspace{-1cm}
 
 \begin{IEEEbiography}[{\includegraphics[width=1in,height=1.25in,clip,keepaspectratio]{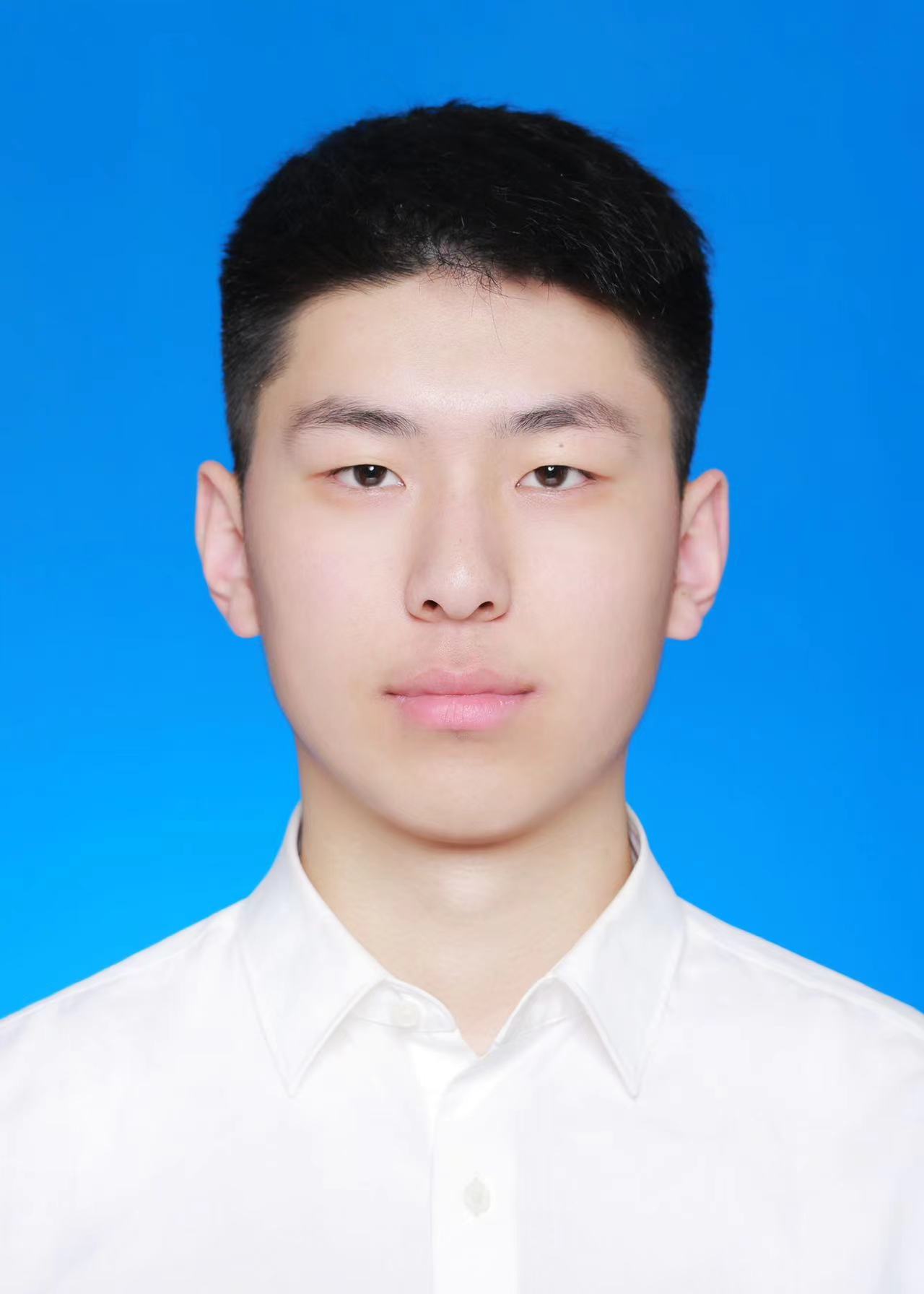}}]{Yuanchang Yue}
        received the B.Sc. degree from jiangnan University, wuxi, China, in 2022. He is currently a master's student with the Aerospace Information Research Institute, Chinese Academy of Sciences. 

        His research interests include embodied intelligence, and task planning. 
	\end{IEEEbiography}
\vspace{-1cm}



 \begin{IEEEbiography}[{\includegraphics[width=1in,height=1.25in,clip,keepaspectratio]{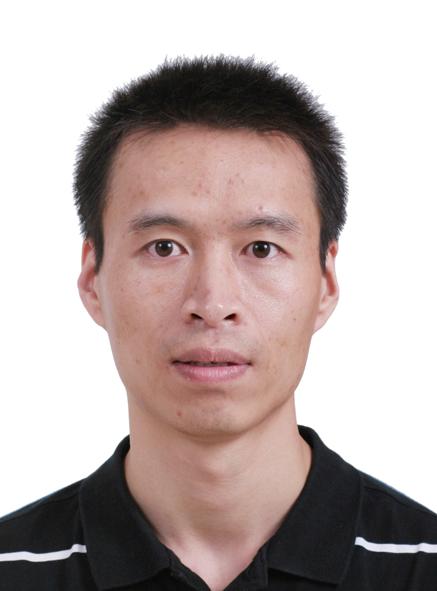}}]{Guangluan Xu}
        received the B.Sc. degree in communications engineering from the Beijing Information Science and Technology University, Beijing, China, in 2000 and the M.Sc. and Ph.D. degrees in signal and information processing from the Institute of Electronics, Chinese Academy of Sciences, Beijing, in 2005.

        He is currently a Professor with the Aerospace Information Research Institute, Chinese Academy of Sciences. His research interests include computer vision and remote sensing image understanding. 
	\end{IEEEbiography}
 \vspace{-1cm}
 
	\begin{IEEEbiography}[{\includegraphics[width=1in,height=1.25in,clip,keepaspectratio]{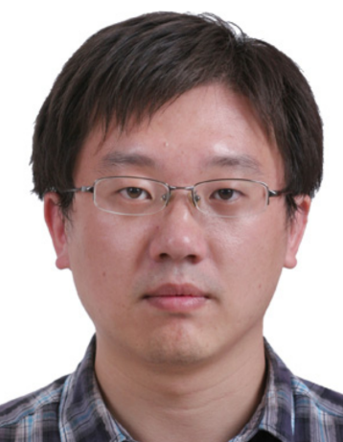}}]{Xian Sun}
		received the B.Sc. degree from the Beijing University of Aeronautics and Astronautics, Beijing, China, in 2004, and the M.Sc. and Ph.D. degrees from the Institute of Electronics, Chinese Academy of Sciences, Beijing, in 2009. 
		
		He is a Professor with the Aerospace Information Research Institute, Chinese Academy of Sciences, Beijing, China. His research interests include computer vision, geospatial data mining, and remote sensing image understanding.
	\end{IEEEbiography}
	\vspace{-1cm}
	
	\begin{IEEEbiography}[{\includegraphics[width=1in,height=1.25in,clip,keepaspectratio]{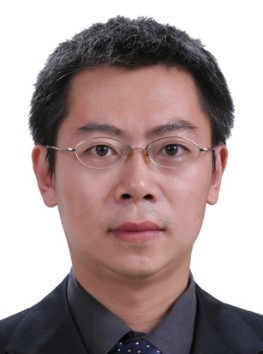}}]{Kun Fu}
		 received the B.Sc., M.Sc., and Ph.D. degrees from the National University of Defense Technology, Changsha, China, in 1995, 1999, and 2002, respectively. 
		 
		 He is a Professor with the Aerospace Information Research Institute, Chinese Academy of Sciences, Beijing, China. His research interests include computer vision, remote sensing image understanding, geospatial data mining, and visualization.
	\end{IEEEbiography}

\newpage



\vfill
\end{document}